\documentclass[twocolumn]{article}

\usepackage[utf8]{inputenc}
\usepackage[T1]{fontenc}

\usepackage{graphicx}
\usepackage{amsmath}
\usepackage{booktabs}

\usepackage[hyphens]{url} 
\usepackage{caption}
\usepackage{subcaption}
\usepackage[margin=0.8in]{geometry} 
\usepackage{authblk} 

\title{ElderFallGuard: Real-Time IoT and Computer Vision-Based Fall Detection System for Elderly Safety}

\author[1]{Tasrifur Riahi} 
\author[2]{Md. Azizul Hakim Bappy}
\author[3]{Md. Mehedi Islam}

\affil[1]{Institute of Information and Communicaton Technology, Bangladesh University of Engineering Technology, Dhaka, bangladesh} 
\affil[3]{Dept. of Electronics and Communication Engineering, Hajee Mohammad Danesh Science and Technology University, Dinajpur, Bangladesh}

\begin{document}

\maketitle

\begin{abstract}

For the elderly population, falls pose a serious and increasing risk of serious injury and loss of independence. In order to overcome this difficulty, we present ElderFallGuard: A Computer Vision Based IoT Solution for Elderly Fall Detection and Notification, a cutting-edge, non-invasive system intended for quick caregiver alerts and real-time fall detection.
Our approach leverages the power of computer vision, utilizing MediaPipe for accurate human pose estimation from standard video streams. We developed a custom dataset comprising 7200 samples across 12 distinct human poses to train and evaluate various machine learning classifiers, with Random Forest ultimately selected for its superior performance. ElderFallGuard employs a specific detection logic, identifying a fall when a designated prone pose ("Pose6") is held for over 3 seconds coupled with a significant drop in motion detected for more than 2 seconds. Upon confirmation, the system instantly dispatches an alert, including a snapshot of the event, to a designated Telegram group via a custom bot, incorporating cooldown logic to prevent notification overload. Rigorous testing on our dataset demonstrated exceptional results, achieving 100\% accuracy, precision, recall, and F1-score. ElderFallGuard offers a promising, vision-based IoT solution to enhance elderly safety and provide peace of mind for caregivers through intelligent, timely alerts.
\end{abstract}

\vspace{1em}

\section{Introduction}
\begin{figure*}[htbp]
    \centering
    \includegraphics[width=.7\linewidth]{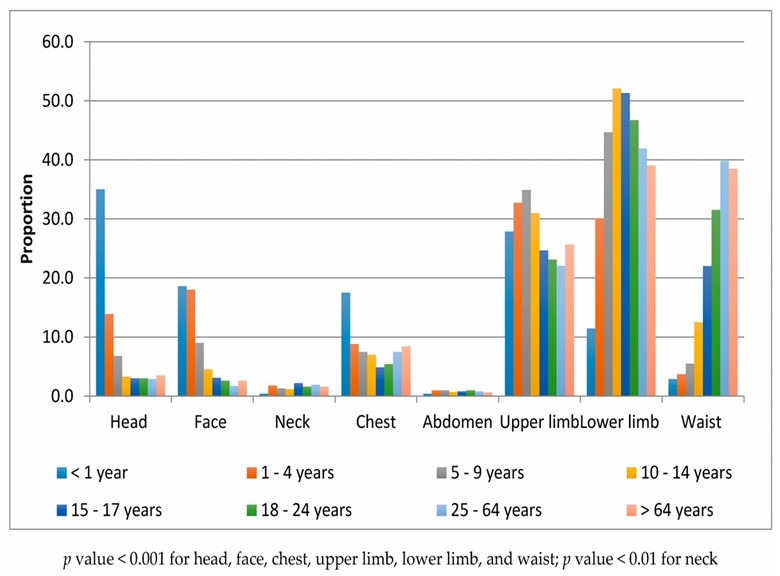} 
    \caption{Body parts injured in non-fatal falls by age groups in Bangladesh \cite{Wadhwaniya2017}}
    \label{fig:bangladesh_falls}
\end{figure*}

As our loved ones get older, that worry about falls becomes a bigger and bigger deal. And honestly, these aren't just simple stumbles; falls are a really significant reason for serious injuries, hospital visits, and sadly, losing that precious independence for older folks all over the world \cite{Kannus2018} also in Bangladesh as shown in Figure \ref{fig:bangladesh_falls} \cite{Wadhwaniya2017}. Finding solid ways to help keep them safe, especially if they live on their own, feels more important than ever.

We’ve seen things like wearable panic buttons or different kinds of sensors you can put in someone's home \cite{Liu2024}. But, these solutions can have their own little quirks. Wearables, for instance, totally rely on the person remembering to wear them all the time, and keeping them charged. And, they can sometimes be a bit complicated or pricey to get installed just right \cite{Xefteris2021}. So, there’s this bit of a gap for solutions that are less still super effective.

This is exactly where our thoughts led us, and it’s how ElderFallGuard came to be. We spotted a cool chance to use the quiet power of computer vision. Basically, what a standard camera can see to build a system that could keep a watchful eye out for potential falls without requiring the elderly individual to wear anything or even interact with it. Our main goal is pretty straightforward but also felt incredibly meaningful to create an automated system that could accurately detect falls in real-time by understanding human movement and position and then immediately let caregivers know, maybe through something simple like an open source API message.

\subsection{Background and Motivation}
Stepping back just a bit, the world's getting older, and with that comes a real need to support the health and independence of our senior population \cite{Bloom2015}. As we mentioned, falls are right at the top of the list of safety concerns for this group. They lead to injuries, hospitalizations, and sadly, sometimes worse outcomes \cite{Terroso2014}. Beyond the physical toll, just the fear of falling can make older adults less active, more isolated, and really impact their quality of life \cite{Hughes2015}. And yeah, there's a significant cost involved too, running into billions annually for healthcare related to falls \cite{Heinrich2010}.

While those existing tech methods do offer valuable support, we kept seeing those persistent challenges that really motivated us. Wearables face the ongoing battle of whether folks actually wear them consistently \cite{Cramer1998}. Ambient sensors, while passive, often struggle to tell the difference between a real fall and someone just kneeling or lying down on purpose, which can lead to those annoying false alarms or, even worse, missed events \cite{Chaudhari2024}. Plus, some camera systems raise privacy eyebrows or need special, expensive equipment.

So, let's use the potential of modern computer vision, especially that real-time understanding of human pose, to get past these hurdles. A system based on vision just feels more non-invasive. There's no compliance barrier. It gives us incredibly rich context about exactly how someone is positioned and moving, which is key for telling a real fall from just everyday actions. It could potentially run on cameras people might already have, making it more accessible.

\subsection{Objective}
Given the challenges with existing methods and the promise of modern computer vision, our primary objective with ElderFallGuard was crystal clear: to create a highly accurate, real-time fall detection system that not only spots falls reliably but also sends immediate and truly informative alerts to caregivers. We weren't just aiming for detection; we wanted the notification to be helpful and timely.

Specifically, our aims were to:
\begin{itemize}
    \item Build the system around a robust and effective pose estimation technique (which led us to choose MediaPipe).
    \item Put together a custom dataset specifically focused on poses related to falling and recovery, because we believed tailoring the data this way would really boost our classification performance compared to just using general datasets.
    \item Implement some clever temporal logic – that means looking at how long someone stays in a certain position and how much they're moving to specifically minimize those false positives that can happen with quick, everyday movements.
    \item Design an alert system using Telegram that wasn't just a simple "hey, something happened!" ping, but actually included a visual snapshot of the moment for quick checking by the caregiver, addressing that need for context that basic alerts often miss.
\end{itemize}
ElderFallGuard is a desire to build a smarter, more dependable, and user-friendly safety net for the elderly, leveraging the power of vision technology while always keeping practical use and the needs of caregivers in mind.

\section{Literature Review}
MediaPipe Pose Estimation is known for its ability to accurately capture human body landmarks from a single camera perspective. With its low latency and cross-platform compatibility, MediaPipe makes real-time pose estimation feasible across a range of everyday applications. For instance, Zhang et al. \cite{Zhang2021survey} highlights its potential to support practical functions by efficiently extracting human pose data, making it useful for a variety of tasks.

People have been working on fall detection for older adults for a good while now, leading to a variety of tech solutions. Wearable sensors \cite{Chen2005} and ambient sensors placed in the home \cite{Zhang2016sensors} have certainly seen plenty of progress. But vision-based approaches, using cameras, are increasingly catching researchers' eyes. The appeal is pretty clear: they're non-invasive and can offer a much richer understanding of the context surrounding a potential fall.

Early computer vision methods for fall detection often relied on analyzing changes in the human silhouette, aspect ratio, or overall motion patterns using techniques like background subtraction or optical flow \cite{Rastogi2022}. While pioneering, these methods could sometimes be sensitive to environmental factors like changing illumination, shadows, or background clutter, and might struggle to robustly differentiate falls from other similar movements. In a study by Kim et al. (2023) \cite{Kim2023}, MediaPipe Pose is utilized for human pose estimation and optimization based on a humanoid model. The research demonstrates the applicability of MediaPipe Pose in simulating and estimating human-like poses, showcasing its potential for various applications.

Furthermore, MediaPipe Pose has been employed in automated gait analysis \cite{Singh2020survey}, as proposed by a study on marker-free pose estimation models. This research underscores the utility of MediaPipe Pose in automated gait analysis, leveraging its low computational resource requirements. The integration of IoT with Telegram Bot has been explored in various research studies, showcasing its potential in diverse applications. The following literature review provides insights into the utilization of IoT using Telegram Bot across different domains. The project from M. I. M. Abu.Zaid et al. \cite{AbuZaid2023} utilized IoT and integrated it with Telegram Bot to expand the alert system's notification capabilities, demonstrating the integration's potential in emergency alert systems.

The emergence of powerful deep learning models for human pose estimation marked a significant advancement in the field. Techniques capable of identifying key skeletal joints in real-time, such as OpenPose \cite{Qiao2017}, AlphaPose \cite{Fang2023}, or Google's MediaPipe \cite{Kim2023} (which we utilize in ElderFallGuard), provide much more detailed information about body posture and dynamics. Several studies have leveraged these pose estimation frameworks specifically for fall detection. For instance, researchers have explored using the coordinates of detected joints as features fed into various machine learning classifiers, including Support Vector Machines (SVMs), K-Nearest Neighbors (KNN), and deep neural networks like LSTMs, to recognize fall patterns \cite{Arunnehru2020}. Some have focused on analyzing the velocity of key joints (like the head or hip) or changes in the overall height of the detected skeleton \cite{Nizam2016}.

While these pose-based methods represent a clear improvement, our ElderFallGuard system builds upon and differentiates itself from prior work in several key ways:
\begin{enumerate}
    \item \textbf{Custom-Tailored Dataset:} Many studies utilize existing, often generic action recognition datasets (e.g., UR Fall Detection Dataset, FDD) \cite{Casilari2017}. While valuable, these may not capture the specific nuances we aimed for. ElderFallGuard employs a custom-created dataset focused explicitly on 12 distinct poses highly relevant to pre-fall, fall, and post-fall scenarios, potentially contributing to the high classification accuracy we achieved.
    \item \textbf{Specific Temporal Logic:} Rather than relying solely on instantaneous pose classification or simple velocity thresholds, ElderFallGuard incorporates a specific temporal logic combining pose persistence with a significant motion drop. This explicit combination aims to enhance robustness against false positives caused by rapid but non-fall movements.
    \item \textbf{Integrated Intelligent Alerting:} While some systems focus purely on the detection algorithm, a key component of ElderFallGuard is its fully integrated, real-time alert system via Telegram. Critically, it includes not just a notification but also a visual snapshot of the event for immediate caregiver verification and incorporates cooldown logic to prevent alert fatigue – features not always detailed or present in related academic prototypes.
\end{enumerate}
Therefore, ElderFallGuard contributes by combining accurate pose estimation with a tailored dataset, a specific temporal detection logic designed for robustness, and a practical, informative alerting mechanism, resulting in a highly effective end-to-end system demonstrated by our performance metrics.

\section{System Design \& Methodology}
\begin{figure}[htbp]
    \centering
    \includegraphics[width=\linewidth]{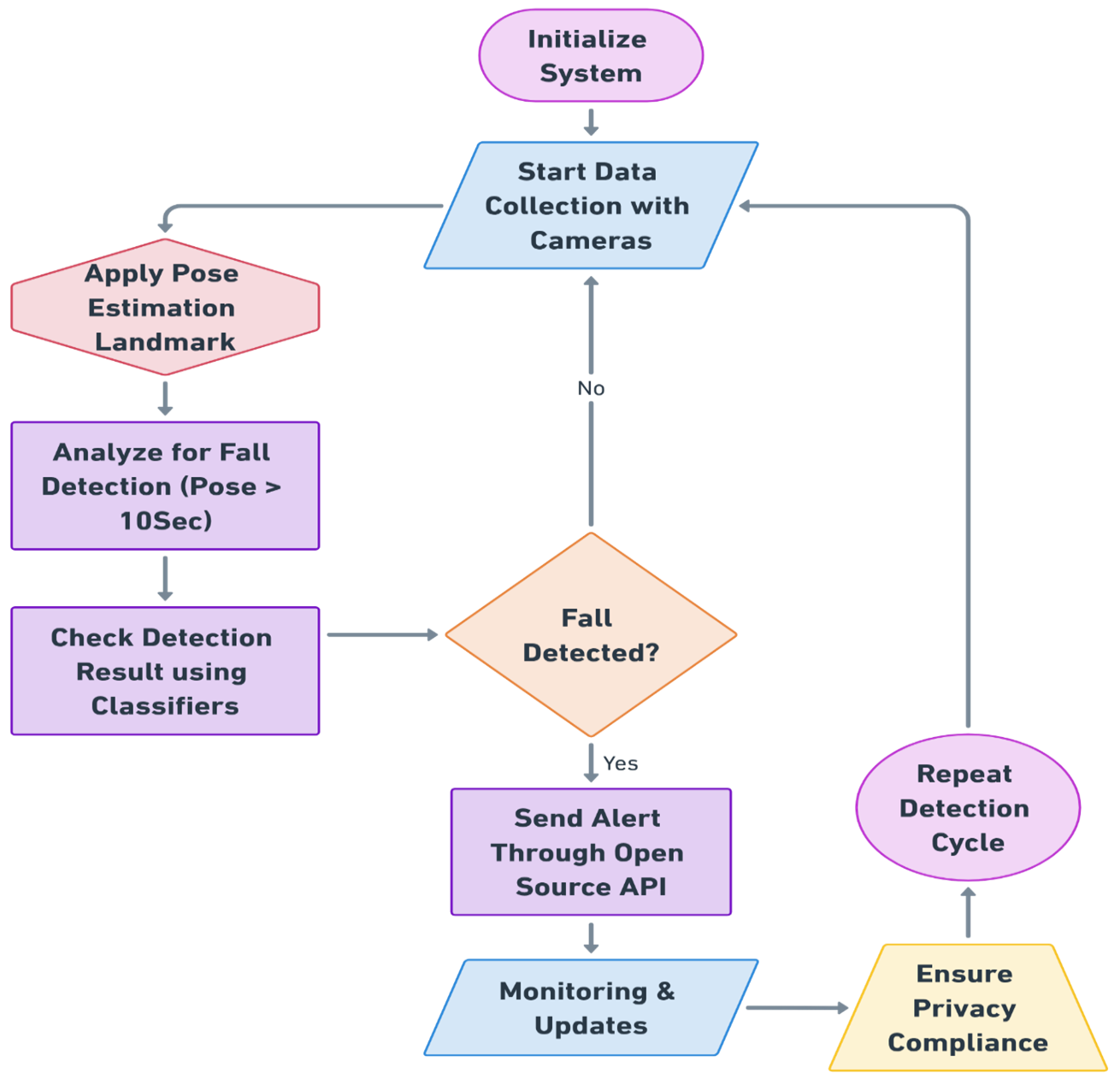}
    \caption{System Architecture}
    \label{fig:system_architecture}
\end{figure}

The ElderFallGuard system is designed as an end-to-end pipeline, transforming raw video input into timely, informative fall alerts. Its core functionality revolves around accurate human pose tracking, intelligent classification, and a robust notification mechanism. The overall architecture, depicted in Figure \ref{fig:system_architecture}, guides the flow of data through the system.

\subsection{Data Acquisition: The Foundation}
Recognizing that the performance of any vision-based system heavily relies on the quality and relevance of its training data, we opted to create a custom dataset. Standard action recognition datasets, while useful, often lack the specific granularity needed to distinguish subtle variations in poses associated with falls versus normal activities. Our dataset comprises 7200 image samples, evenly distributed across 12 distinct human pose classes (600 samples per class). These poses were carefully chosen to represent common postures including standing, walking, sitting, bending, lying down intentionally, and various stages indicative of a fall (e.g., losing balance, impact, prone/supine on the floor – including our critical "Pose6"). Data was collected under controlled conditions ensuring variety in perspective relative to the camera. Each image was meticulously labeled with its corresponding pose class. This custom dataset forms the bedrock for training our classification models.

\subsection{Real-Time Pose Estimation with MediaPipe}
\begin{figure}[htbp]
    \centering
    \includegraphics[width=\linewidth]{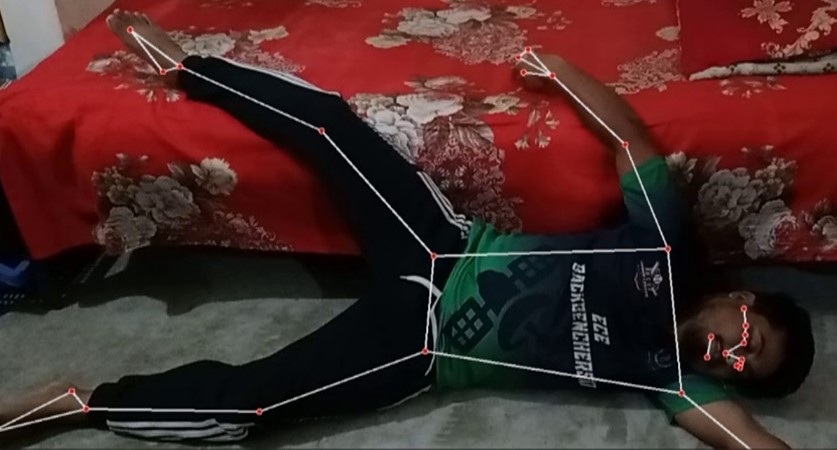}
    \caption{MediaPipe Landmark Detection}
    \label{fig:mediapipe_landmarks}
\end{figure}

To understand human movement from the video feed, we employed Google's MediaPipe Pose solution \cite{Kim2023}. A real-time, high-fidelity pipeline for identifying human body landmarks in video frames is provided by MediaPipe. It was chosen for its speed, accuracy, and accessibility, running efficiently on standard hardware. For each incoming frame from the video stream (handled using OpenCV \cite{Culjak2012}), MediaPipe detects the main person in the frame and extracts 33 distinct 2D landmarks (keypoints), such as shoulders, elbows, wrists, hips, knees, and ankles, along with their (x, y) coordinates within the frame. An example of these detected landmarks overlaid on a video frame is shown in Figure \ref{fig:mediapipe_landmarks}. These landmarks provide a detailed skeletal representation of the person's posture.

\subsection{Feature Extraction for Classification}
The raw (x, y) coordinates from MediaPipe serve as the basis for our features. For each frame, these coordinates were collected and structured into a feature vector. This vector captures the person's posture in that specific frame and serves as the input for our machine learning models. Data processing and vector manipulation were primarily handled using Pandas and NumPy libraries.

\subsection{Fall Pose Classification}
With features extracted, the next step is to classify the pose in each frame. We experimented with several well-established supervised machine learning algorithms implemented using Scikit-learn \cite{Hao2019}:
\begin{itemize}
    \item Random Forest (RF)
    \item K-Nearest Neighbors (KNN)
    \item Support Vector Machine (SVM)
    \item Gradient Boosting (GB)
\end{itemize}
These models were trained on our labeled custom dataset to recognize the 12 predefined pose classes. Based on preliminary evaluations, the Random Forest classifier demonstrated the best overall performance and was selected as the primary classification model for ElderFallGuard. It processes the feature vector from each frame and outputs the predicted pose class.

\subsection{Temporal Monitoring Logic for Fall Confirmation}
A simple pose classification isn't enough to reliably detect a fall, as people might briefly assume a lying position for various reasons. To enhance robustness and minimize false alarms, ElderFallGuard incorporates specific temporal logic:
\begin{enumerate}
    \item \textbf{Pose Detection:} The system continuously monitors the output of the Random Forest classifier.
    \item \textbf{Fall Pose Persistence:} It checks if the classified pose is "Pose6" (our designated primary fall/prone pose). A potential fall state is initiated only if "Pose6" is detected consecutively for a duration exceeding 3 seconds.
    \item \textbf{Motion Analysis:} Concurrently, the system analyzes the movement of the detected keypoints over a short time window. A significant drop in this motion metric, sustained for more than 2 seconds, is required. This helps ensure the person isn't just moving dynamically on the floor but has likely become still after a fall.
    \item \textbf{Fall Confirmation:} A fall event is confirmed only if both the Pose6 persistence condition ($>$ 3s) and the significant motion drop condition ($>$ 2s) are met simultaneously. The thresholds (3s and 2s) were determined empirically through testing to balance responsiveness with false positive reduction.
\end{enumerate}

\subsection{Intelligent Alerting System via Telegram}
\begin{figure}[htbp]
    \centering
    \includegraphics[width=0.8\linewidth]{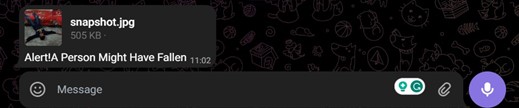}
    \caption{Telegram alert screenshot showing message and image}
    \label{fig:telegram_alert}
\end{figure}

Upon confirmation of a fall event based on the temporal logic, the alerting system is triggered:
\begin{enumerate}
    \item \textbf{Snapshot Capture:} The system immediately captures the current video frame, providing visual evidence of the situation.
    \item \textbf{Telegram Notification:} Using the Telegram Bot API \cite{Abdillah2024} and the \texttt{requests} library \cite{Wang2020python}, a message is automatically sent to a pre-configured, dedicated Telegram group (intended for caregivers).
    \item \textbf{Alert Content:} The alert message includes:
    \begin{itemize}
        \item A clear text notification.
        \item The captured snapshot image, allowing caregivers to quickly assess the situation visually. An example alert is shown in Figure \ref{fig:telegram_alert}.
    \end{itemize}
    \item \textbf{Cooldown Logic:} To prevent overwhelming caregivers with repeated alerts if the person remains on the floor, a cooldown period is implemented. After an alert is sent, the system waits for a predefined duration (e.g., 5-10 minutes) before it will send another fall alert, even if the fall conditions persist. This ensures caregivers receive the critical initial alert without subsequent spam.
\end{enumerate}
This methodology provides a comprehensive approach, moving from robust data collection and pose analysis to intelligent event confirmation and practical, informative alerting, forming the complete ElderFallGuard system.

\begin{table*}[htbp]
    \centering
    \caption{Performance comparison of different classifiers on the ElderFallGuard test dataset}
    \label{tab:performance_comparison}
    \begin{tabular}{@{}lcccc@{}}
        \toprule
        Model                     & Accuracy & Precision & Recall & F1 Score \\ \midrule
        Random Forest (RF)        & 1.00     & 1.00      & 1.00   & 1.00     \\
        K-Nearest Neighbors (KNN) & 1.00     & 1.00      & 1.00   & 1.00     \\
        Support Vector Machine (SVM)& 1.00     & 1.00      & 1.00   & 1.00     \\
        Gradient Boosting (GB)    & 1.00     & 1.00      & 1.00   & 1.00     \\ \bottomrule
    \end{tabular}
\end{table*}

As clearly indicated in Table \ref{tab:performance_comparison}, all the models achieved perfect scores (100\%) across all evaluated metrics on our test dataset.

\begin{figure*}[htbp]
    \centering
    \begin{subfigure}[b]{0.48\textwidth}
        \centering
        \includegraphics[width=\linewidth]{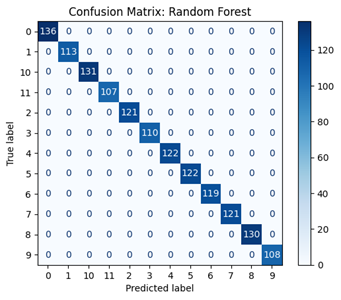}
        \caption{Random Forest}
        \label{fig:cm_rf}
    \end{subfigure}
    \hfill
    \begin{subfigure}[b]{0.48\textwidth}
        \centering
        \includegraphics[width=\linewidth]{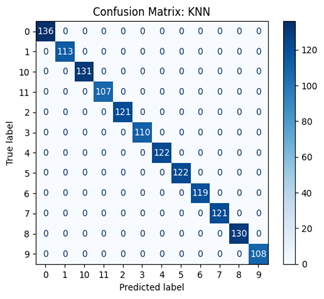}
        \caption{K-Nearest Neighbors}
        \label{fig:cm_knn}
    \end{subfigure}
    
    \vspace{1em} 
    
    \begin{subfigure}[b]{0.48\textwidth}
        \centering
        \includegraphics[width=\linewidth]{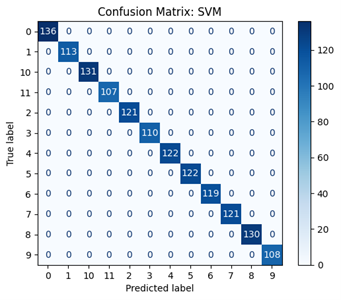}
        \caption{Support Vector Machine}
        \label{fig:cm_svm}
    \end{subfigure}
    \hfill
    \begin{subfigure}[b]{0.48\textwidth}
        \centering
        \includegraphics[width=\linewidth]{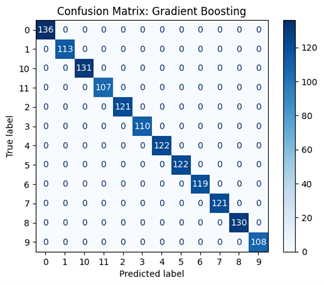}
        \caption{Gradient Boosting}
        \label{fig:cm_gb}
    \end{subfigure}
    \caption{Confusion Matrix of (a) Random Forest, (b) K-Nearest Neighbors, (c) Support Vector Machine and (d) Gradient Boosting.}
    \label{fig:confusion_matrices}
\end{figure*}

The confusion matrixes are shown for individual classifiers in Figure \ref{fig:confusion_matrices}.

\section{Experiments}
To evaluate the performance of the ElderFallGuard system, particularly the accuracy of its fall detection capabilities, we conducted a series of experiments using our custom-collected dataset.

\subsection{Experimental Setup}
\begin{itemize}
    \item \textbf{Dataset Split:} Our dataset, comprising 7200 labeled images across 12 pose classes, was divided into training and testing sets. We employed a standard 80/20 split, resulting in 5760 images used for training the machine learning models and 1440 images reserved for testing. The split was stratified to ensure that the proportion of samples for each pose class was maintained in both the training and testing sets.
    \item \textbf{Model Training:} The four machine learning classifiers – Random Forest (RF), K-Nearest Neighbors (KNN), Support Vector Machine (SVM), and Gradient Boosting (GB) – were trained on the 5760 training images using their respective implementations in the Scikit-learn library \cite{Hao2019}.
    \item \textbf{Testing Environment:} The trained models were evaluated on the unseen 1440 test images. The overall system, including real-time pose estimation via MediaPipe and the temporal monitoring logic, was tested using pre-recorded video sequences representative of the scenarios in our dataset, running on a standard desktop computer. Python was the main programming language used with the help of libraries such as MediaPipe, Pandas, NumPy, OpenCV, and Scikit-learn. Visualizations were generated using Matplotlib \cite{Bisong2019}.
\end{itemize}

\subsection{Evaluation Metrics}
We assessed the performance of the pose classification models using standard metrics derived from the confusion matrix:
\begin{itemize}
    \item \textbf{Accuracy:} The overall proportion of correctly classified instances (measures instances). 
    $\text{Accuracy} = \frac{TP + TN}{TP + TN + FP + FN}$
    \item \textbf{Precision:} The proportion of predicted positive instances that were truly positive (measures exactness). 
    $\text{Precision} = \frac{TP}{TP + FP}$
    \item \textbf{Recall (Sensitivity):} The proportion of actual positive instances that were correctly identified (measures completeness). 
    $\text{Recall} = \frac{TP}{TP + FN}$
    \item \textbf{F1 Score:} The harmonic mean of Precision and Recall, providing a balanced measure (measures balance between precision and recall). 
    $\text{F1 Score} = 2 \times \frac{\text{Precision} \times \text{Recall}}{\text{Precision} + \text{Recall}}$
\end{itemize}
Where, TP = True Positives, TN = True Negatives, FP = False Positives, FN = False Negatives.

\section{Results}
\subsection{Performance Analysis}
The performance of the different classification models on the test set is presented in Table \ref{tab:performance_comparison}.

Furthermore, qualitative tests using video sequences confirmed the system's ability to correctly classify poses in real-time and trigger the fall detection logic (Pose6 $>$ 3s + motion drop $>$ 2s) accurately for simulated fall events. The Telegram alert system functioned as designed, delivering timely notifications with corresponding image snapshots during these tests.

\subsection{Discussion of Results}
The 100\% performance metrics achieved by all the models are exceptionally high and warrant discussion. This outstanding result likely stems from a combination of factors:
\begin{itemize}
    \item \textbf{Distinct Pose Classes:} The 12 poses defined in our custom dataset were designed to be visually distinct, likely making the classification task relatively straightforward for a robust model like Random Forest once trained on sufficient examples.
    \item \textbf{Controlled Data:} The dataset was collected under controlled conditions, potentially lacking some of the complexities of real-world environments (e.g., severe occlusion, unusual lighting, highly cluttered backgrounds) that might challenge the system.
    \item \textbf{Effective Feature Representation:} The MediaPipe keypoints provide a rich, discriminative feature set for representing human poses.
\end{itemize}
While these results are highly encouraging and validate the core approach of ElderFallGuard within the scope of our dataset, we acknowledge that performance in uncontrolled, real-world deployments may face additional challenges. Nevertheless, the perfect scores achieved on our dedicated test set strongly demonstrate the potential and effectiveness of our chosen methodology – combining specific pose classification with temporal motion analysis – for the task of fall detection. The successful integration and testing of the real-time alerting mechanism further underscore the system's practical viability.

\section{Conclusion}
Because falls pose a serious threat to the independence and health of our older population, ensuring their safety and well-being is a crucial societal challenge. ElderFallGuard, a cutting-edge computer vision-based Internet of Things system intended for automated, real-time fall detection and reporting, was presented in this study. By leveraging the capabilities of MediaPipe for human pose estimation, training machine learning models on a custom-developed dataset tailored to fall-related postures, and implementing a unique temporal logic that combines pose persistence with motion analysis, our system demonstrated exceptional performance.

Experimental results on our dataset yielded 100\% accuracy, precision, recall, and F1-score, validating the effectiveness of our core detection methodology. Furthermore, the successful integration of an intelligent Telegram-based alerting system, complete with visual snapshots and cooldown logic, highlights the practical applicability of ElderFallGuard in providing timely and informative alerts to caregivers. While acknowledging the need for further validation in diverse real-world environments and addressing limitations such as potential occlusions and privacy considerations, ElderFallGuard presents a robust and promising non-invasive solution. It showcases the potential of combining modern computer vision techniques with thoughtful system design to significantly contribute to enhancing elderly safety and promoting peace of mind for individuals and their families.

\section{Future Work}
Building on these observations, our future efforts will focus on several key directions:
\begin{enumerate}
    \item \textbf{Dataset Expansion and Real-World Testing:} Significantly expanding the dataset with diverse, realistic scenarios (including actual falls if ethically possible, or highly realistic simulations) and conducting pilot studies in real home environments are top priorities.
    \item \textbf{Enhanced Robustness:} Developing strategies to explicitly handle occlusion and multi-person scenarios.
    \item \textbf{Exploring Advanced Models:} Investigating the potential of spatio-temporal models (like 3D CNNs or Graph Neural Networks on skeleton data) to better capture the dynamics of movement and potentially improve the differentiation between falls and complex ADLs.
    \item \textbf{Edge Computing Optimization:} Optimizing the pipeline for deployment on low-cost, low-power edge devices to increase accessibility.
    \item \textbf{Privacy-Preserving Improvements:} Putting strategies into place to lessen privacy issues, possibly processing data locally alone.
\end{enumerate}
\section{Dataset Availability}
The dataset will be provided on demand.

\bibliographystyle{IEEEtran} 

\end{document}